\documentclass[letterpaper, 10 pt, conference]{ieeeconf}  % Comment this line out if you need a4paper

\IEEEoverridecommandlockouts                              % This command is only needed if 
\usepackage{graphics} % for pdf, bitmapped graphics files
\usepackage{amsmath} % assumes amsmath package installed
\usepackage{amssymb}  % assumes amsmath package installed
\usepackage{amsfonts}
\usepackage{cite}
\usepackage{algorithm}
\usepackage{algorithmic}
\usepackage{graphicx}
\usepackage{subfigure }
\usepackage{multirow}
\usepackage{makecell}
\usepackage{color}

\newcommand{\vect}[1]{{\boldsymbol{#1}}}

\newcommand{\E}{\mathbb{E} }
\newcommand{\R}{\mathbb{R} }
\newcommand{\KL}{D_{\textrm{KL}}}
\newcommand{\blue}[1]{{\color{black}#1}}

\title{\LARGE \bf
Robustifying a Policy in Multi-Agent RL with Diverse Cooperative Behaviors and Adversarial Style Sampling for Assistive Tasks
}

\author{Takayuki Osa$^{1,2}$, Tatsuya Harada$^{1,2}$ % <-this % stops a space
\thanks{$^{1}$T. Osa and T. Harada are with Graduate School of Information Science and Technology, the University of Tokyo, Tokyo, Japan
        {\tt\small \{osa, harada\}@mi.t.u-tokyo.ac.jp}}%
\thanks{$^{2}$T. Osa and T. Harada are also with RIKEN Center for Advanced Intelligence Project (AIP), Tokyo, Japan.}%
}

\begin{document}

\maketitle
\thispagestyle{empty}
\pagestyle{empty}

%%%%%%%%%%%%%%%%%%%%%%%%%%%%%%%%%%%%%%%%%%%%%%%%%%%%%%%%%%%%%%%%%%%%%%%%%%%%%%%%
\begin{abstract}
Autonomous assistance of people with motor impairments is one of the most promising applications of autonomous robotic systems. 
Recent studies have reported encouraging results using deep reinforcement learning (RL) in the healthcare domain.
Previous studies showed that assistive tasks can be formulated as multi-agent RL, wherein there are two agents: a caregiver and a care-receiver.
However, policies trained in multi-agent RL are often sensitive to  the policies of other agents. 
In such a case, a trained caregiver's policy may not work for different care-receivers. 
To alleviate this issue, we propose a framework that learns a robust caregiver's policy by training it for diverse care-receiver responses. 
In our framework, diverse care-receiver responses are autonomously learned through trials and errors. 
In addition, to robustify the care-giver's policy, we propose a strategy for sampling a care-receiver's response in an adversarial manner during the training.
We evaluated the proposed method using tasks in an Assistive Gym. 
We demonstrate that policies trained with a popular deep RL method are vulnerable to changes in policies of other agents and that the proposed framework improves the robustness against such changes.
\end{abstract}

%%%%%%%%%%%%%%%%%%%%%%%%%%%%%%%%%%%%%%%%%%%%%%%%%%%%%%%%%%%%%%%%%%%%%%%%%%%%%%%%
\section{Introduction}

In the United states, it was reported that approximately 26\% of adults have some type of disability, and 3.7\% of the 26\% that have a form of disability have difficulty in self-care, including behavior such as dressing and bathing~\cite{Okoro18}. 
To assist such people with motor impairments, assistive robotics systems have been investigated for decades~\cite{Chen13}.
Recent advances in machine learning and robotics have developed quickly, and recent studies have demonstrated impressive results for various applications.
Reinforcement  learning~(RL)~\cite{Sutton18}, which is an approach for learning the optimal behavior through autonomous trials and errors, has been applied to a diverse range of applications, including robotic control~\cite{Levine16b} and autonomous driving\cite{Kendall19}.
However, although many advancements have been made in RL research, there still exist many challenges in general~\cite{Ibarz21}, and assistive robots in particular.

\begin{figure}
	\centering
	 \subfigure{\includegraphics[width=0.46\columnwidth]{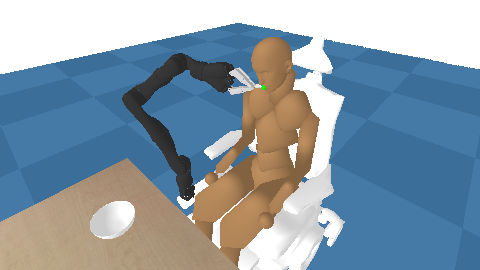}}
	\subfigure{\includegraphics[width=0.46\columnwidth]{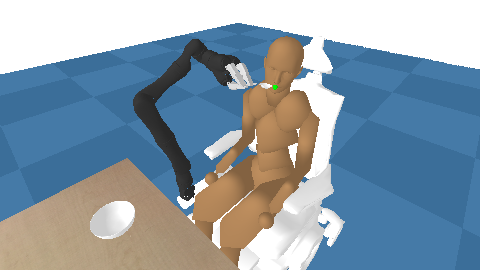}}
	\subfigure{\includegraphics[width=0.46\columnwidth]{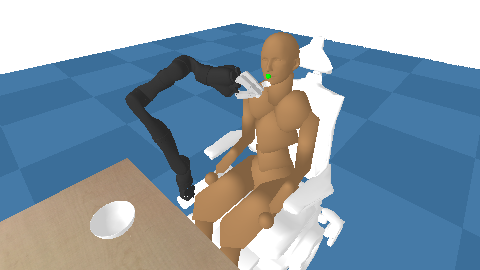}}
	\subfigure{\includegraphics[width=0.46\columnwidth]{jaco_z09-09step35.png}}
%	\vskip -0.1in
	\caption{Diverse care receiver's responses for the feeding task in Assistive Gym~\cite{Erickson20}. 
    Our framework autonomously learns diverse care receiver's responses and robustifies the caregiver's policy in an adversarial-training fashion. }
%	\vskip -0.15in
	\label{fig:intro}
\end{figure}

Previous studies in~\cite{Erickson20,Clegg20} developed a simulator for assistive robots and showed that assistive tasks can be formulated as multi-agent RL, wherein there are two agents: a caregiver and a care-receiver. 
In their framework, a deep RL method was applied to both agents, and each agent learned natural behavior through autonomous trial and error during the simulation. 
This framework was referred to as co-optimization~\cite{Erickson20,Clegg20}.
Alternatively, a limitation of this framework is that the performance of the caregiver's policy depends on the care-receiver's policy, 
and therefore  the caregiver's policy may not work if the care-receiver's policy is changed.
This vulnerability of a policy is a common problem in multi-agent RL.
In the literature of multi-agent RL, it is reported that the performance of the learned policies is often highly sensitive to the policies of other agents~\cite{Li17}.
In practice, the behavior of the care-receiver is unknown and diverse in the real world.
When we transfer a caregiver's policy trained in simulation to the real world, the behavior of a care-receiver must be different from that obtained with co-optimization in simulation.
Therefore, it is essential to consider the robustness of the policy against the change in the care-receiver's policy.

To address this issue, we propose a framework that robustifies the caregiver's policy by learning diverse behaviors of the care-receiver in co-optimization.
Our contribution is to propose a practical algorithm for learning a caregiver's policy that is robust against the change in the care-receiver's behavior for assistive tasks.
By training the caregiver's policy for diverse care-receiver's responses, we can ensure that the caregiver's policy is robust against the changes in the care-receiver's behavior.
In prior work, diverse care-receiver's responses were obtained by manually designing various reward functions for the care-receiver~\cite{He22}. 
By contrast, our framework does not require such reward engineering because diverse care-receiver's responses is autonomously obtained by maximizing the mutual information.
Furthermore, we propose to sample the care-receiver's behavior style in an adversarial-training fashion during training.
Even if we train the caregiver's policy for diverse care-receiver's responses, we observed that uniformly sampling the diverse care-receiver's response did not lead to satisfactory performance of the caregiver's policy.
Our strategy for sampling the caregiver's response style can be considered as adversarial training that improves the worst case performance.
We evaluated the proposed method using tasks in Assistive Gym~\cite{Erickson20}.
The experimental results shows that caregiver's policies obtained by a standard co-optimization are actually vulnerable to the change in the care-receiver's policy.
Our results demonstrate that the caregiver's policy obtained by the proposed framework is more robust against changes in the care-receiver's policy compared with that obtained by co-optimization using a widely-used deep RL method.

\section{Related Work}
Assistive robots have been investigated as a promising approach for empowering people with motor impairments~\cite{Chen13}.
Previous studies addressed the application of assistive robots to tasks such as dressing~\cite{Zhang19,Clegg20} and feeding~\cite{Rhodes18,Park18,Gallenberger19}.
Within the context of learning assistive tasks, imitation learning~\cite{Osa18} and reinforcement learning~\cite{Sutton18} are two of the most popular approaches.
However, a limitation of imitation learning is that collecting expert demonstrations is time-consuming and thousands of demonstrations are often necessary to obtain satisfactory generalization performance~\cite{Osa18}.
In contrast, the optimal policy is learned through autonomous trials and errors in RL~\cite{Sutton18}.
RL has been successful in various domains, including robotic manipulation~\cite{Ibarz21}, board games~\cite{Schrittwieser20}, and autonomous driving~\cite{Kiran21}.
Unlike imitation learning, it is not necessary to provide human demonstrations, and the performance of RL agents often outperforms human experts~\cite{Silver17}.
However, performing trials and errors in real robotic systems can be costly and risky in practice.
To address this issue,  Erickson et al. recently developed a simulator, Assistive Gym, which is designed for assistive tasks to accelerate the research in this field~\cite{Erickson20}.
Assistive Gym encompasses various tasks, including dressing, drinking, and feeding.
Our work on multi-agent RL for assistive tasks is built on top of Assistive Gym.

Based on tasks in Assistive Gym, Clegg et al. showed that natural assistive motions can be obtained by jointly optimizing policies for a caregiver and care-receiver~\cite{Clegg20}.
However, in their framework, the resulting policy of the caregiver is dependent on the care-receiver's policy and thus may not work for a different care-receiver who have a different behavior style.
Recent work by He et al. addressed this issue from a meta-learning approach~\cite{He22}.
They prepared diverse care-receiver's response by manually engineering the reward function for the care-receiver in co-optimization.
Subsequently, the latent space of the care-receiver's responses are learned.
The caregiver's policy was adapted to the care-receiver's response by estimating the corresponding latent variable through the interaction.
Although the approach proposed in~\cite{He22} is promising, it requires manual engineering of the reward function and running co-optimization multiple times to obtain diverse care-receiver's responses.
In contrast, our framework does not require such reward function engineering to obtain diverse care-receiver's response styles because we leverage an algorithm based on mutual information maximization~\cite{Osa22}.
Furthermore, diverse care-receiver's responses are learned by performing co-optimization just one time in our framework.

\section{Background}
\subsection{Assistive Tasks as Multi-agent Reinforcement Learning}
In RL, we consider a Markov decision process~(MDP) that consists of a tuple~$(\mathcal{S}, \mathcal{A}, \mathcal{P}, r  , \gamma, d)$ 
where $\mathcal{S}$ is the state space,  $\mathcal{A}$ is the action space,  $\mathcal{P}(\vect{s}_{t+1}|\vect{s}_t, \vect{a}_t)$ is the transition probability density,  
$r(\vect{s}, \vect{a})$ is the reward function, $\gamma$ is the discount factor, and $d(\vect{s}_0)$ is the probability density of the initial state.
A policy $\pi(\vect{a}|\vect{s}): \mathcal{S} \times \mathcal{A} \mapsto \R$ is defined as the conditional probability density over actions given the states.
The aim of RL is to learn a policy that maximizes the expected return $\E[R_0|\pi] $ where $R_t = \sum_{k=t}^{T} \gamma^{k-t}r(\vect{s}_k, \vect{a}_k)$.

In this work, we are particularly interested in a setting in which there are two agents, namely a caregiver and care-receiver, in assistive tasks.
Multi-agent RL is often formulated based on Markov games, which is an extension of MDPs~\cite{Littman94}.
A Markov game for $N$ agents is defined by a state space $\mathcal{S}$, a set of action spaces for $N$ agents $\mathcal{A}_1, \ldots, \mathcal{A}_N$, and a set of observation spaces for $N$ agents $\mathcal{O}_1, \ldots, \mathcal{O}_N$.
In our problem setting, the agent $i$ cannot observe the action taken by another agent $j$ for $j \neq i$.
This problem setting is referred to as a partially observable Markov game~\cite{Littman94,Lowe17}.
In multi-agent RL, each agent $i$ tries to maximize the its own expected returns $\E[R^i_0 | \pi^i]$.

In our problem setting, the two agents share the same reward function and aim to achieve successful and comfortable care. 
Therefore, the two agents \textit{cooperate} to achieve the same goal~\cite{Lowe17,Ackermann19}.
In this study, we propose a framework for learning diverse behaviors in a cooperative multi-agent RL setting.
We denote the caregiver's action and state by $\vect{a}_{\textrm{g}} \in \mathcal{A}_{\textrm{g}}$ and $\vect{s}_{\textrm{g}} \in \mathcal{S}_{\textrm{g}}$, respectively.
Similarly, we denote the care-receiver's action and state by $\vect{a}_{\textrm{r}} \in \mathcal{A}_{\textrm{r}}$ and $\vect{s}_{\textrm{r}} \in \mathcal{S}_{\textrm{r}}$, respectively.
We also denote the policies of the caregiver and care-receiver by $\pi^{\textrm{g}}$ and $\pi^{\textrm{r}}$, respectively.

\subsection{Latent-conditioned policies for modeling diverse behaviors}
Previous studies~\cite{Eysenbach19,Kumar20,Sharma20} have showed that diverse behaviors can be represented by a policy conditioned on a latent variable $\pi(\vect{a}|\vect{s}, \vect{z})$, where $\vect{z}$ is the latent variable.
They proposed methods for training the latent-conditioned policy such that it changes the output according to the value of the latent variable $\vect{z}$.
When we have a latent-conditioned policy such that it changes the output according to the value of the latent variable $\vect{z}$,
sampling a policy from a distribution over policies can be approximated with sampling a value of latent variable $\vect{z}$. 
%In this framework, when latent variable $\vect{z}$ is continuous, policy $\pi$ is given by
%\begin{align}
%	\pi(\vect{a}|\vect{s}) = \int p(\vect{z}) \pi(\vect{a}|\vect{s}, \vect{z}) d\vect{z},
%\end{align}
%where $p(\vect{z})$ is the prior distribution of latent variable $\vect{z}$.
The value of the latent variable is sampled at the beginning of an episode and fixed until the end of the episode. 
%The distribution of the trajectory conditioned on the latent variable is given by
%\begin{align}
%	p(\vect{\tau} | \vect{z}) = d(\vect{s}_0) \prod_{t=0}^{T} \pi(\vect{a}_t | \vect{s}_t, \vect{z}) \mathcal{P}(\vect{s}_{t+1} | \vect{s}_t, \vect{a}_t).
%\end{align}
In this framework, latent-conditioned policy $\pi(\vect{a}|\vect{s}, \vect{z})$ is evaluated based on the state function conditioned on the latent variable defined by
\begin{align}
	V^\pi(\vect{s}, \vect{z}) = \E_{\pi}[ R |\vect{s}, \vect{z} ],
\end{align}
which represents the expected return when starting in state $\vect{s}$ and following policy $\pi$ given the latent variable $\vect{z}$.

\section{Robusfitying Policies in Assistive Tasks}
In this section, we first consider how to train the caregiver's policy for diverse care-receiver's response.
Subsequently, we present how to obtain diverse care-receiver's responses during training.
Then, we describe the adversarial style sampling to robustify the caregiver's policy.

% \subsection{Adversarial Style Sampling for Training a Robust Caregiver's Policy}
\subsection{Training Caregiver's Policy for Diverse Care-Receiver's responses}

% \subsubsection{Expected Return for a Set of Policies of a Caregiver and a Care-Receiver} 
In our framework, we aim to obtain a robust caregiver's policy by training it with diverse care-receiver's behavior. 
%We denote by $\pi^{\textrm{g}}_{\vect{\theta}}$ the caregiver's policy parameterized with a vector $\vect{\theta}$.
%Similarly, we denote by $\pi^{\textrm{r}}_{\vect{\phi}}$ the caregiver's policy parameterized with a vector $\vect{\phi}$.
Assuming that the distribution of the care-receiver's policy $\pi^{\textrm{r}}$ is given by $p(\pi^{\textrm{r}})$,
the following expected return should be maximized to perform the task appropriately:
\begin{align}
\max_{\pi^g, \pi^r} \E_{\pi^{\textrm{r}} \sim p(\pi^{\textrm{r}})} \left[ \E_{\vect{s}' \sim \mathcal{P}}[R|\pi^{\textrm{g}}, \pi^{\textrm{r}}] \right].
	\label{eq:return}
\end{align}
We approximate this problem using the latent-conditioned policy, which we introduced in the previous section. 
As we can specify the type of the behavior of the latent-conditioned policy by setting the value of the latent variable, we can approximate the expected return in \eqref{eq:return} with 
\begin{align}
   \max_{\pi^g, \pi^r} \E_{\vect{z}_{\textrm{r}} \sim p(\vect{z}_{\textrm{r}})} \left[ \E_{\vect{s} \sim \mathcal{P}}[R|\pi^{\textrm{g}}(\vect{a}_{\textrm{g}}|\vect{s}_{\textrm{g}}), \pi^{\textrm{r}}(\vect{a}_{\textrm{r}}|\vect{s}_{\textrm{r}}, \vect{z}_{\textrm{r}})] \right],
	\label{eq:return2}
\end{align}
where $\vect{z}_{\textrm{r}}$ is the latent variable that specifies the behavior of the care-receiver, respectively, and  $p(\vect{z}_{\textrm{r}})$ is the prior distributions of $\vect{z}_{\textrm{r}}$. 
The expectation in \eqref{eq:return2} can be approximated using samples stored in the replay buffer in an off-the-shelf RL algorithm.
In this study, we use PPO as a base RL algorithm~\cite{Schulman17}, although other RL algorithms can also be used.
We denote by by $\vect{\mu}_{\vect{\theta}}(\vect{s}_{\textrm{g}})$ the caregiver's deterministic policy parameterized with a vector $\vect{\theta}$.
Once we collect samples through the interaction between the caregiver's policy and diverse care-receiver's responses, we update the caregiver's policy to maximize the expected return.
%by maximizing the following objective function:
%\begin{align}
%    \mathcal{J}(\vect{\theta}) =  \sum_{ \vect{s}_{\textrm{g}}\in \mathcal{B}_{\textrm{g}}}   Q^{\textrm{g}}_{\vect{w}}(\vect{s}_{\textrm{g}}, \vect{\mu}_{\vect{\theta}}(\vect{s}_{\textrm{g}}) ),
%	\label{eq:ltd3_care_giver}
%\end{align}
%where $Q^{\textrm{g}}_{\vect{w}}(\vect{s}_{\textrm{g}},\vect{a}_{\textrm{g}})$ is the Q-function corresponding to the caregiver's policy, and  $\mathcal{B}_{\textrm{g}}$ is a batch of samples selected from a caregiver's replay buffer $\mathcal{D}_{\textrm{g}}$, comprising samples stored through the interaction between the caregiver's policy and diverse care-receiver's responses.

\subsection{Learning diverse care-receiver's responses}
\label{sec:div_receiver}
To learn diverse behaviors of the care-receiver, we extend PPO~\cite{Schulman17} to a method that trains a latent-conditioned policy by maximizing the mutual information between the latent variable and the state-action pairs.
In the context of multi-agent RL, it is reported that training each agent using PPO often leads the performance better than state-of-the-art multi-agent RL methods~\cite{Yu21}.
% Maximizing the mutual information to obtain diverse behavior is a prevalent approach in deep RL~\cite{Eysenbach19,Kumar20}.
% Recently it is reported that LTD3 can learn diverse behaviors for tasks where there are multiple solutions~\cite{Osa22}.
To train a latent-conditioned policy, we consider the problem of maximizing the following objective function, which can be obtained by extending the one in \cite{Schulman17}.
\begin{align}
\mathcal{L}_{\textrm{adv}}(\vect{\theta}) = \E_{\vect{z}, \vect{s}_{\textrm{r}}, \vect{a}_{\textrm{r}} \sim p(\vect{z}), d^{\pi_{\textrm{old}}}, \pi_{\textrm{old}}} \left[ L_{\textrm{clip}}(\vect{s}_{\textrm{r}}, \vect{a}_{\textrm{r}}, \vect{z})\right],
\end{align}
subject to
\begin{align}
\E_{\vect{s}\sim d^\pi} \left[ \KL\left( \pi^{\textrm{r}}_{\textrm{old}}(\vect{a}|\vect{s}, \vect{z})||\pi^{\textrm{r}}_{\vect{\theta}}(\vect{a}|\vect{s},\vect{z})  \right) \right] < \eta,
\label{eq:constraint_lppo}
\end{align}
where 
\begin{align}
L_{\textrm{clip}}(\vect{s}_{\textrm{r}}, \vect{a}_{\textrm{r}}, \vect{z}) & = \min \left( r(\vect{\theta}) A_t, \tilde{r}(\vect{\theta}) A_{\textrm{r}}\right), \\
r(\vect{\theta}) &= \frac{\pi^{\textrm{r}}_{\vect{\theta}}(\vect{a}|\vect{s},\vect{z})}{\pi^{\textrm{r}}_{\textrm{old}}(\vect{a}|\vect{s},\vect{z})}, \\
\tilde{r}(\vect{\theta})& = \textrm{clip}(r(\vect{\theta}), 1 - c, 1 + c),
\end{align}
$c$ is a constant, and $A_{\textrm{r}}$ is the advantage function for the care-receiver's policy.
The constraint in \eqref{eq:constraint_lppo} constrains the change of the policy conditioned on the latent variable stabilizes the learning process.
While solving the above problem leads to obtain a latent-conditioned policy that maximizes the expected return, the diversity of the behaviors encoded in the policy is not encouraged.

To encourage the diversity of the behaviors, we maximize the lower bound of the mutual information between the latent variable and the state-action pairs induced by policy $\pi$, which we denote by $I_{\pi}(\vect{z};\vect{s}, \vect{a})$.
As shown in \cite{Osa22}, the variational lower bound of $I_{\pi}(\vect{z};\vect{s}, \vect{a})$ is given by
\begin{align}
 I_{\pi}(\vect{z};\vect{s}, \vect{a}) \geq \E_{(\vect{s}, \vect{a}) \sim \pi, \mathcal{P}} [ \log q_{\vect{\phi}}(\vect{z}|\vect{s}, \vect{a}) ],
	\label{eq:lower_bound_im}
\end{align}
where $q_{\vect{\phi}}(\vect{z}|\vect{s}, \vect{a})$ is an approximated posterior distribution parameterized with a vector $\vect{\phi}$.
As the right-hand side of \eqref{eq:lower_bound_im} is a simple log-likelihood, maximizing this lower bound is tractable in practice.  
Based on \eqref{eq:lower_bound_im}, we train the care-receiver's policy maximizing the following objective function:
\begin{align}
\mathcal{L}_{\textrm{LPPO}}(\vect{\theta}) = \mathcal{L}_{\textrm{adv}}(\vect{\theta}) + \alpha \E_{\vect{s}_{\textrm{r}}, \vect{a}_{\textrm{r}} \sim \pi^{\textrm{r}}, \mathcal{P}} [ \log q_{\vect{\phi}}(\vect{z}_{\textrm{r}}|\vect{s}_{\textrm{r}}, \vect{a}_{\textrm{r}}) ]
\label{eq:lppo_care_receiver}
\end{align}
subject to
\begin{align}
	\E_{\vect{s}\sim d^\pi} \left[ \KL\left( \pi_{\textrm{old}}(\vect{a}|\vect{s}, \vect{z})||\pi_{\vect{\theta}}(\vect{a}|\vect{s},\vect{z})  \right) \right] < \eta,
\end{align}
where $\alpha$ is a constant that balance the weight on the expected return and mutual information terms.
We refer to the resulting algorithm as Latent-conditioned Proximal Policy Optimization (LPPO).

\subsection{Adversarial Style Sampling}
When we design a practical algorithm to solve the problem in \eqref{eq:return2}, the choice of $p(\vect{z}_{\textrm{r}})$ is crucial to obtain a satisfactory performance of the caregiver's policy.
In the context of learning diver solutions in RL, a popular choice is to use the uniform distribution as $p(\vect{z}_{\textrm{r}})$~\cite{Kumar20}.
If we use the uniform distribution as $p(\vect{z}_{\textrm{r}})$, the average performance over the diverse care-receiver's responses is maximized.
Although maximizing the average performance is reasonable, the worst-case performance is not considered in this case.
As indicated in the literature of risk-aware RL~\cite{Bodnar20}, it is often necessary to improve the worst-case performance to obtain a robust policy.

Based on the above consideration, we propose to solve the following max-min problem: 
\begin{align}
   \max_{\pi^g, \pi^r} \min_{\tilde{p}(\vect{z}_{\textrm{r}})} \E_{\vect{z}_{\textrm{r}} \sim \tilde{p}(\vect{z}_{\textrm{r}})} \left[ \E_{\vect{s} \sim \mathcal{P}}[R|\pi^{\textrm{g}}(\vect{a}_{\textrm{g}}|\vect{s}_{\textrm{g}}), \pi^{\textrm{r}}(\vect{a}_{\textrm{r}}|\vect{s}_{\textrm{r}}, \vect{z}_{\textrm{r}})] \right],
\label{eq:return_adv}
\end{align}
which can be viewed as a type of adversarial training in which the caregiver's and care-receiver's policies $\pi^g$, $\pi^r$ are updated so as to maximize the expected rewards, and the behavior style sampler $\tilde{p}(\vect{z}_{\textrm{r}})$ is set to minimize the expected reward during training.
In this problem formulation, the agent attempts to maximize the worst-case performance, leading to improve the robustness of the caregiver's policy.

To solve the problem in \eqref{eq:return_adv}, we sample the latent variable as follows during training:
\begin{align}
	\vect{z}_{\textrm{r}} = \arg \min_{\tilde{\vect{z}}_{\textrm{r}}} \E_{\vect{s} \sim \mathcal{P}}[R|\pi^{\textrm{g}}(\vect{a}_{\textrm{g}}|\vect{s}_{\textrm{g}}), \pi^{\textrm{r}}(\vect{a}_{\textrm{r}}|\vect{s}_{\textrm{r}}, \tilde{\vect{z}}_{\textrm{r}})].
\end{align}
As the value of the latent variable specifies the behavior style of the care-receiver, this approach enable us to select the behavior style of the care-receiver that lead to the worst performance during the training.
In our framework, we consider cooperative tasks where the reward is shared between the caregiver and care-receiver.
Therefore, the approximated latent-conditioned state value, $V^{\textrm{r}}_{\vect{w}}(\vect{s}_{\textrm{r}}, \vect{z}_{\textrm{r}})$,
indicates the expected return when the care-receiver takes the response corresponding to the value of $\vect{z}_{\textrm{r}}$ under state $\vect{s}_{\textrm{r}}$.
In our implementation, the value of the latent variable $\vect{z}_{\textrm{r}}$ is sampled as follows:
\begin{align}
\vect{z}_{\textrm{r}} = \arg \min_{\vect{z}_{\textrm{r}}}\frac{1}{N}\sum_{(\vect{s}_{\textrm{r}}, \vect{z}_{\textrm{r}})\in \mathcal{B}}   V^{\textrm{r}}_{\vect{w}}(\vect{s}_{\textrm{r}}, \vect{z}_{\textrm{r}}).
\label{eq:adv_sampling}
\end{align}
When latent variable $\vect{z}_{\textrm{r}}$ is continuous, it is not feasible to analytically perform the minimization in \eqref{eq:adv_sampling}.
In practice, we generate $M$ samples of latent variable $\vect{z}_{\textrm{r}}$ from the uniform distribution $U(-1, 1)$, and use the sample with the lowest value as the minimizer.
Meanwhile, to encourage the diversity of the care-receiver's response, it is also necessary to sample a wide range of values of $\vect{z}_{\textrm{r}}$ during training.
Thus, in practice, we sample the value of $\vect{z}_{\textrm{r}}$ in a $\epsilon$-greedy-like fashion;
with probability $\epsilon$, the value of $\vect{z}_{\textrm{r}}$ is determined by \eqref{eq:adv_sampling}. Otherwise, the value of $\vect{z}_{\textrm{r}}$ is sampled from the uniform distribution $U(-1, 1)$.
% This $\epsilon$-greedy like strategy can be seen as maximizing the sum of the objective functions in \eqref{eq:return_adv} and \eqref{eq:return2} where $p(\vect{z})$ is given by the uniform distribution.

\subsection{Practical algorithm}

\begin{algorithm}[tb]
	\caption{Robustifying the caregiver's policy with diverse care-receiver's response and adversarial style sampling}
	\label{alg:ltd3_multi}
	\begin{algorithmic}
		\STATE {\bfseries Input:} Dimension of latent variable $\vect{z}_{\textrm{r}}$ for the care-receiver's policy, $\epsilon$ for adversarial style sampling
		\STATE Initialize policies $\pi_{\textrm{g}}$, $\pi_{\textrm{r}}$ and buffers $\mathcal{D}_{\textrm{g}}$, $\mathcal{D}_{\textrm{r}}$
		\REPEAT
		\WHILE{the data size in the buffers is not sufficient}
        \STATE sample $x_{\textrm{rng}}$ from the uniform distribution $U(0,1)$
        \IF{ $x_{\textrm{rng}} < \epsilon$}
		\STATE Set the latent variable $\vect{z}_{\textrm{r}}$ with \eqref{eq:adv_sampling}
        \ELSE
        \STATE Sample $\vect{z}_{\textrm{r}}$ from the uniform distribution $U(-1,1)$
        \ENDIF
		\FOR{$t=0$ {\bfseries to} $T$}
		\STATE  Select actions with exploration noise for each agent
		\STATE Observe reward $r$ and new state $ \vect{s}'_{\textrm{g}}$ and  $\vect{s}'_{\textrm{r}}$
    	\STATE Store tuple $(\vect{s}_{\textrm{g}}, \vect{a}_{\textrm{g}}, \vect{s}'_{\textrm{g}}, r)$ in $\mathcal{D}_{\textrm{g}}$
    	\STATE Store tuple $(\vect{s}_{\textrm{r}}, \vect{a}_{\textrm{r}}, \vect{s}'_{\textrm{r}}, r, \vect{z}_{\textrm{r}})$ in $\mathcal{D}_{\textrm{r}}$
%    	\STATE Sample mini-batch $\mathcal{B}_i$ from $\mathcal{D}_i$
    		% \STATE Update the critics with Double Clipped Q-learning
		\ENDFOR
		\ENDWHILE
		\STATE Update the care-receiver's policy by maximizing $\mathcal{L}_{\textrm{LPPO}}(\vect{\phi})$ in \eqref{eq:lppo_care_receiver} 
		\STATE Update the caregiver's policy with PPO
		\STATE Empty the buffers $\mathcal{D}_{\textrm{g}}$ and $\mathcal{D}_{\textrm{r}}$
		\UNTIL{ $\pi_{\textrm{g}}$ and $\pi_{\textrm{r}}$ are optimized}
	\end{algorithmic}
\end{algorithm}

Our algorithm is summarized in Algorithm~\ref{alg:ltd3_multi}.
The caregiver and care-receiver retain separate replay buffers for each.
Latent variable $\vect{z}_{\textrm{r}}$ that specifies the care-receiver's behavior style is stored in the care-receiver's replay buffer.
The value of the latent variable $\vect{z}_{\textrm{r}}$ is sampled at the beginning of an episode and fixed until the end of the episode.
We set $\epsilon=0.5$ for the adversarial style sampling in our implementation.
When training the care-receiver's policy, we set $\alpha=0.2$ in \eqref{eq:lppo_care_receiver}. 
%As in a standard TD3, the caregiver's policy is updated once every two time steps. 
%Similarly, the care-receiver's policy is updated by maximizing $\mathcal{J}_{Q}(\vect{\phi})$ in \eqref{eq:ltd3_care_receiver_Q} once every two time steps.
%In addition, the parameters $\vect{\phi}$ and $\vect{\psi}$ are udpated once every five time steps by maximizing $\mathcal{J}_{\textrm{info}}(\vect{\phi}, \vect{\psi})$ in \eqref{eq:ltd3_care_receiver_info}.
%While it is assumed that TD3 is used to train the caregiver's policy in Algorithm~\ref{alg:ltd3_multi}, other algorithms such as PPO can also be used to train the caregiver's policy.
%Similarly, although LTD3 is used to learn diverse behaviors in Algorithm~\ref{alg:ltd3_multi}, the mutual information maximization for learning diverse behaviors can also be applied to other algorithms.

\section{Evaluation}
In the experiment, we investigated the following points: 1) sample-efficiency of the propose method in the training, and 2) robustness of the caregiver's policy against the change in the care-receiver's policy.

\begin{figure}
	 	\vspace{-0.25cm}
	\centering
	\subfigure[FeedingPR2Human-v1.]{\includegraphics[width=0.49\columnwidth]{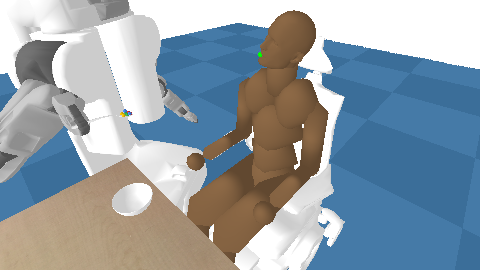}}
	\subfigure[FeedingJacoHuman-v1.]{\includegraphics[width=0.49\columnwidth]{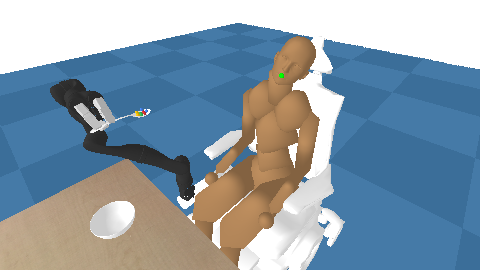}}
	\subfigure[FeedingBaxterHuman-v1.]{\includegraphics[width=0.49\columnwidth]{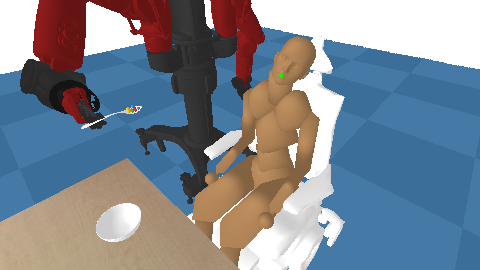}}
	\subfigure[FeedingSawyerHuman-v1.]{\includegraphics[width=0.49\columnwidth]{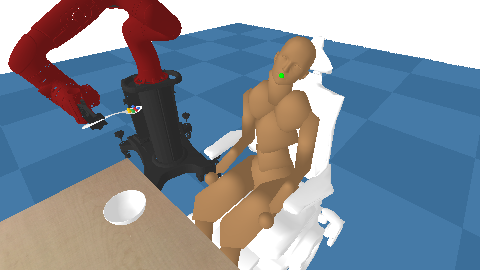}}
	\caption{Tasks in Assistive Gym used in the evaluation.}
	\vskip -0.2in
	\label{fig:task}
\end{figure}

We evaluated the proposed method using tasks implemented in Assistive Gym~\cite{Erickson20}, which is based on the PyBullet physics engine~\cite{coumans2021}.
We used FeedingPR2Human-v1, FeedingJacoHuman-v1, FeedingBaxterHuman-v1, and FeedingSawyerHuman-v1 in our evaluation, as shown in Fig.~\ref{fig:task}.
Each episode consists of 200 time steps. 
%The actions of the caregiver (robot) are represented as changes in the joint positions, and the robot is position-controlled during the simulation. 
%The actions of the care-receiver (human) are represented as a motion of the head, which is four-dimensional. 
Observations of the caregiver include the position and orientation of the spoon and head of the care-receiver and the joint angles of the caregiver. 
Observations of the care-receiver include the position and orientation of the spoon and head of the care-receiver and the joint angles of the care-receiver. 
%The reward function is carefully designed to incorporate natural human preferences, as described in \cite{Erickson20}. 
The reward function is computed based on the distance between the spoon and the mouth, the state of the food (e.g., spilled or not), and the velocity of the end-effector.
For more details on Assistive Gym, please refer to \cite{Erickson20}.

\begin{table}[tb]
	\caption{Hyperparameters for PPO/LPPO}
	\vskip -0.15in
	\label{tbl:hyperparam_ppo}
	\begin{center}
		\begin{tabular}{|l|c|c|}
			\hline
			Parameter & Value & Method \\
			\hline
			Optimizer    & Adam & PPO/LPPO\\
			Policy learning rate & $3 \cdot 10^{-4}$ & PPO/LPPO\\
			Critic learning rate & $1 \cdot 10^{-3}$ & PPO/LPPO\\
			Discount factor $\gamma$ & 0.99 & PPO/LPPO\\
			Steps per epoch & $4000$ & PPO/LPPO\\
			Number of hidden layers & 2 & PPO/LPPO\\
			Number of hidden units & (64, 64) & PPO\\
			Number of hidden units & (128, 64) & LPPO\\
			Activation function & tanh & PPO/LPPO\\
			Coefficient for GAE $\lambda$  & 0.95 & PPO/LPPO\\
			Clip ratio $c$ & 0.2 & PPO/LPPO\\
			Target KL & 0.01 & PPO/LPPO\\
			$\alpha$ & 0.2 & LPPO \\ 
			\hline
		\end{tabular}
	\end{center}
	\vskip -0.35in
\end{table}

To investigate the effect of learning diverse care-receiver's response and the adversarial style sampling, we evaluated the following methods.
As a baseline, we evaluate policy performance in the case where both caregiver's and care-receiver's policies were trained with PPO.
We refer to this baseline as \textit{PPO-PPO}.
Similarly, we refer to the case where both caregiver's and care-receiver's policies were trained with TD3~\cite{Fujimoto18} as \textit{TD3-TD3}.
TD3-TD3 and PPO-PPO can be considered as baselines based on a standard co-optimization proposed in~\cite{Erickson20}. 
To investigate the effect of learning diverse care-receiver's responses, we evaluated policy performance in the case where the caregiver's policy was trained with PPO and the care-receiver's policy with LPPO as described in Section~\ref{sec:div_receiver}. 
we refer to this variant of the proposed method as \textit{PPO-LPPO}.
The differences in policy performance between PPO-PPO and PPO-LPPO indicates the effect of learning diverse care-receiver's responses during training.
Finally, we evaluated the proposed method that incorporates the adversarial style sampling with PPO-LPPO, which is referred to as \textit{PPO-LPPO-adv}.
To investigate the effect of the algorithm for learning diverse behaviors, we also evaluated variants using LTD3~\cite{Osa22}, which is an existing method for learning diverse behaviors in RL. 
TD3-LTD3 refers to the method where the caregiver's policy was trained with TD3 and the care-receiver's policy with LTD3. 
Similarly, TD3-LTD3-adv refers to the method that incorporates the proposed adversarial style sampling with TD3-LTD3.
%We used a two-dimensional continuous latent variable for LPPO.
For LPPO and LTD3, the latent variable of the care-receiver's policy was two-dimensional.
In LPPO and LTD3, we used the uniform distribution $U(-1, 1)$ as the prior distribution of the latent variable $p(\vect{z})$.
The implementation of PPO and TD3 were adapted from spinningup~\cite{spinningup}.
Hyperparameters of PPO and LPPO are summarized in Table~\ref{tbl:hyperparam_ppo}.

% \subsection{Diversity of Behaviors Obtained by the Proposed Method}

\subsection{Learning Curve}
\begin{figure}
	\centering
	\subfigure[FeedingPR2Human-v1.]{\includegraphics[width=0.49\columnwidth]{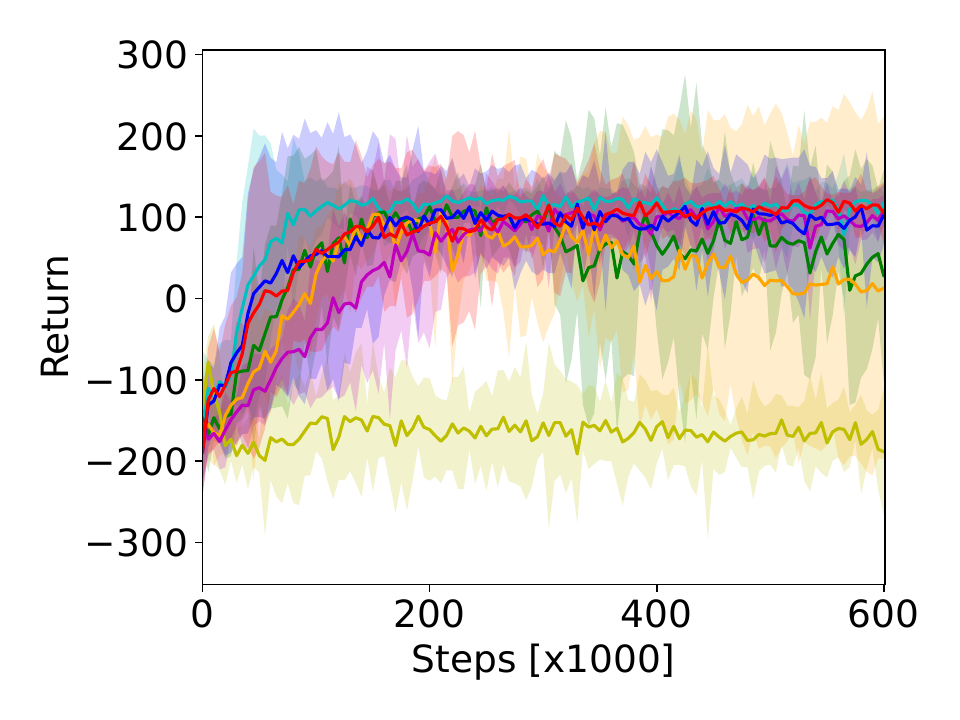}}
	\subfigure[FeedingJacoHuman-v1.]{\includegraphics[width=0.49\columnwidth]{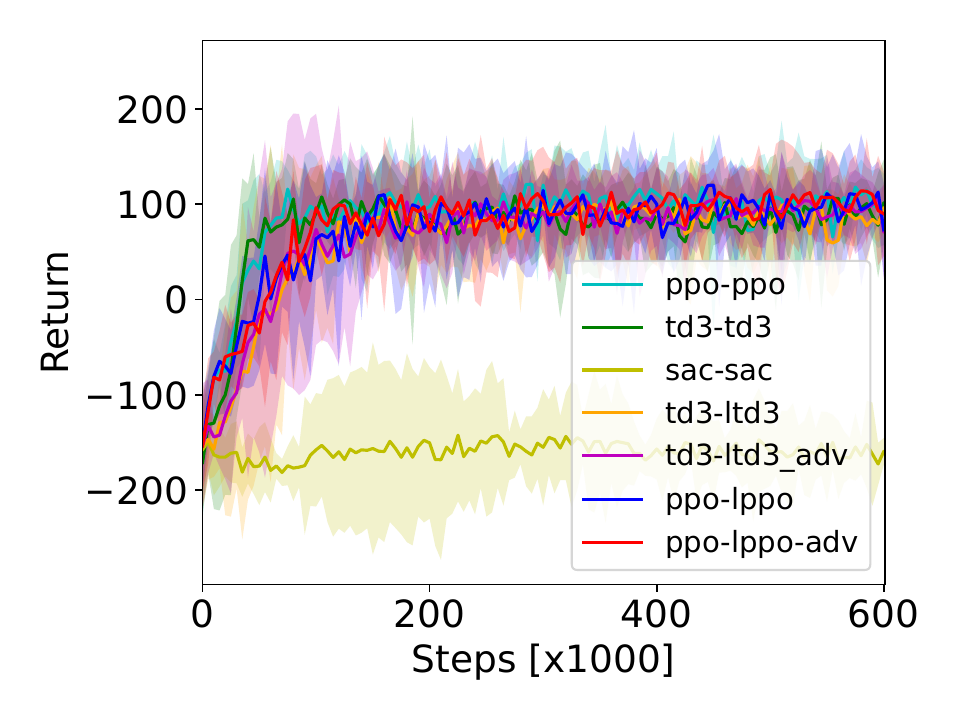}}
	\subfigure[FeedingBaxterHuman-v1.]{\includegraphics[width=0.49\columnwidth]{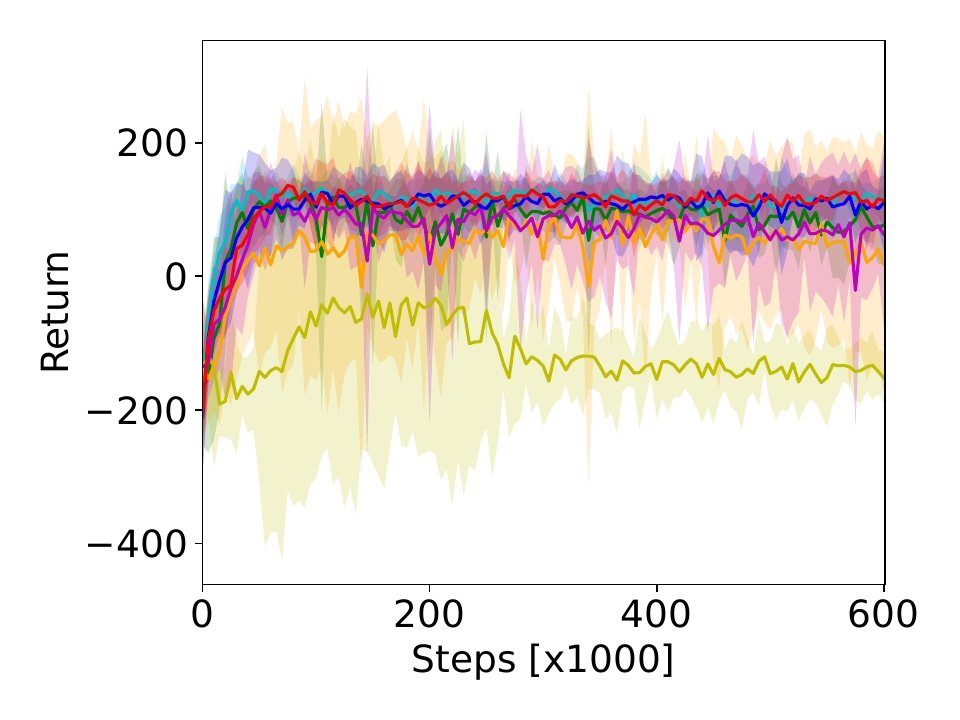}}
	\subfigure[FeedingSawyerHuman-v1.]{\includegraphics[width=0.49\columnwidth]{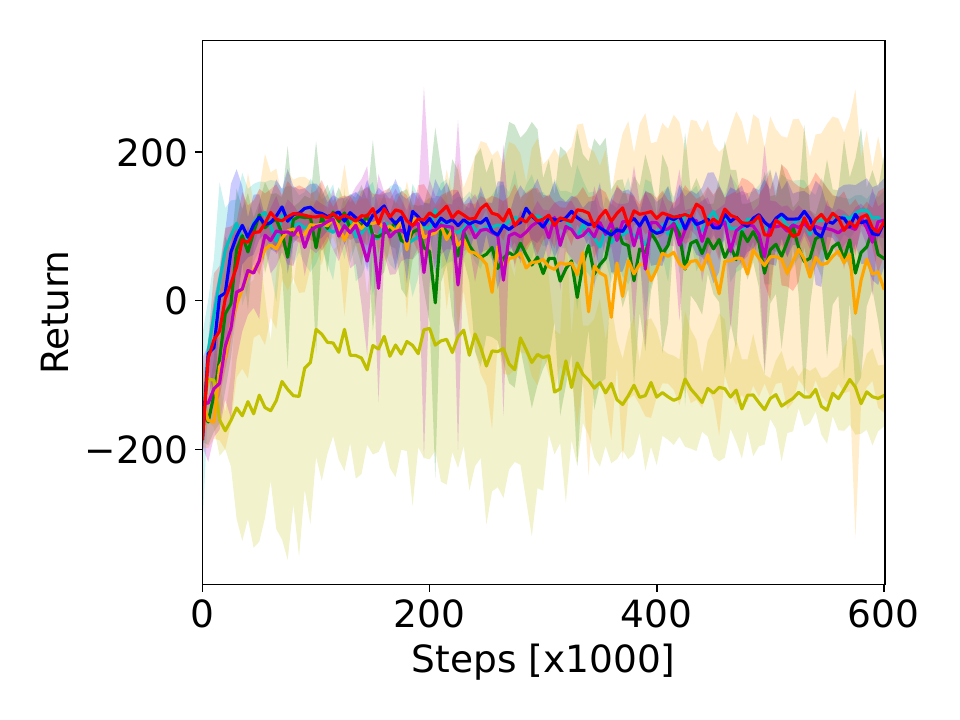}}
		\vskip -0.1in
	\caption{Learning curves of the proposed and baseline methods.}
	\label{fig:learing_curve}
\end{figure}

The learning curves during the traing are shown in Fig.~\ref{fig:learing_curve}.
Regarding FeedingPR2Human-v1, while the performance of TD3-TD3 and TD3-LTD3 often dropped after 4 million steps, 
the performances of PPO-LPPO-adv, PPO-LPPO and PPO-PPO were stable during training.
The difference between LTD3-based methods and LPPO-based methods implies that LPPO-based methods are more stable in cooperative multi-agent RL.
Interestingly, there was no significant difference in the performance and sample-efficiency among PPO-LPPO-adv, PPO-LPPO and PPO-PPO. 
This result indicates that learning diverse care-receiver's responses in PPO-LPPO, and PPO-LPPO-adv does not have a significant effect on the sample-efficiency of the training in these tasks,
although LPPO learns diverse behaviors of the care-receiver.

 \begin{figure}
% 	 	\vspace{-0.5cm}
 	\centering
 	\subfigure[]{\includegraphics[width=0.49\columnwidth]{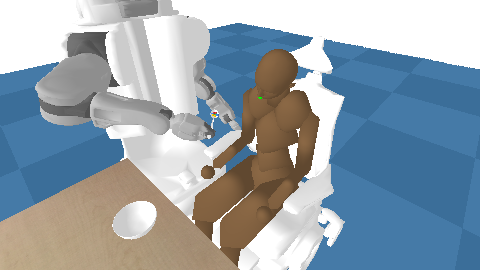}}
 	\subfigure[]{\includegraphics[width=0.49\columnwidth]{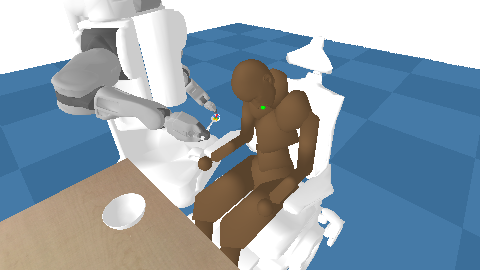}}
 	\subfigure[]{\includegraphics[width=0.49\columnwidth]{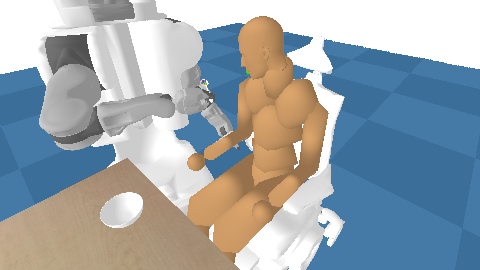}}
 	\subfigure[]{\includegraphics[width=0.49\columnwidth]{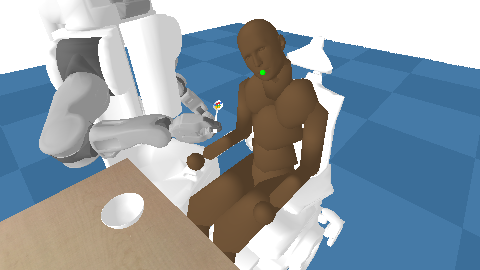}}
 	\vskip -0.1in
 	\caption{Diverse behaviors of the care-receiver obtained for the FeedingPR2Human-v1 task.  The orientation of the care-receiver's head changes according to the value of the latent variable.  (a)-(d) correspond to \blue{$\vect{z}_r=[0.9,0.9], [-0.9,0.9], [0.9,-0.9], [-0.9,-0.9]$}, respectively. Human size and color are randomly set.  }
 	\label{fig:human_diversity}
 	\vspace{-0.3cm}
 \end{figure}

We visualized the behavior of policies obtained in PPO-LPPO-adv after training with 4 million time steps.
We obtained diverse behaviors of the care-receiver in PPO-LPPO-adv, as shown in Fig.~\ref{fig:human_diversity}.
%When the value of the latent variable continuously changed, the behavior of the care-receiver also changed continuously.
As shown in Fig.~\ref{fig:human_diversity}, the care-receiver cooperated with the caregiver by moving the head towards the spoon in different ways.
This result implies that there can be diverse behaviors of the care-receiver and that the caregiver's policy should be robust against the diversity of the care-receiver's behavior.
It is worth noting that it is not trivial to hard-code the reward function to obtain diverse care-receiver's responses shown in Fig.~\ref{fig:human_diversity}. 
The diversity of the care-receiver's responses shown in Fig.~\ref{fig:human_diversity} supports the validity of our framework based on mutual information maximization for learning diverse behaviors.
%In addition, the caregiver's policy successfully performed a feeding task for the variations shown in Fig.~\ref{fig:human_diversity}.
%The results also indicate that we can collect the experiences with diverse care-receivers' responses by training the care-receiver's policy using LTD3 in our multi-agent RL setting.

\begin{table*}[tb]
	\caption{Returns in collaboration with the care-receiver trained separately with TD3}
		\vskip -0.15in
	\label{tbl:td3_test}
	\begin{center}
		\begin{tabular}{|c||c|c|c|c|c|c|c|c|}
			\hline
			\multirow{2}{*}{Methods } & \multicolumn{2}{c|}{FeedingPR2Human-v1} &  \multicolumn{2}{c|}{FeedingJacoHuman-v1}  &  \multicolumn{2}{c|}{FeedingBaxterHuman-v1}  &  \multicolumn{2}{c|}{FeedingSawyerHuman-v1} \\\cline{2-9}
			%			& \makecell{Care-giver and \\ receiver trained \\ together}  & \makecell{Care-receiver \\ trained separately \\ with TD3} & \makecell{Care-giver and \\ receiver trained \\ together}  & \makecell{Care-receiver \\ trained separately \\ with TD3} & \makecell{Care-giver and \\ receiver trained \\ together} & \makecell{Care-receiver \\ trained separately \\ with TD3}\\
			&  training  & test & training  & test & training & test& training & test \\
			\hline
			\makecell[l]{PPO-LPPO-adv (ours) } & \textbf{114.3$\pm$36.1} & \textbf{77.1$\pm$67.8}  & \textbf{101.0$\pm$71.0} & \textbf{89.1$\pm$82.4}  & \textbf{120.5$\pm$40.7} & \textbf{104.7$\pm$47.9} & \textbf{111.7$\pm$54.6} & 87.8$\pm$68.3\\
			\makecell[l]{PPO-LPPO (ours)} & 98.4$\pm$57.4 & \textbf{87.8$\pm$62.4}  & \textbf{97.5$\pm$80.5} & \textbf{84.6$\pm$92.2}  & 103.9$\pm$50.7 & \textbf{103.0$\pm$63.4} & \textbf{115.9$\pm$44.3} & 79.1$\pm$79.1\\
			\makecell[l]{PPO-PPO} & \textbf{114.9$\pm$33.4} & 47.0$\pm$81.8 & 107.7$\pm$71.6 & 81.7$\pm$84.8 & 105.9$\pm$48.4 & 60.3$\pm$73.1 & 107.1$\pm$51.9 & 49.1$\pm$95.9 \\
			\makecell[l]{TD3-LTD3-adv (ours)} & \textbf{112.8$\pm$29.4} & 74.1$\pm$67.6  & 77.4$\pm$69.6 & \textbf{91.6$\pm$60.3}  & 68.7$\pm$70.4 & 41.4$\pm$100.0 & 92.9$\pm$54.0 & \textbf{100.0$\pm$49.8}\\
			\makecell[l]{TD3-LTD3 }  & 88.8$\pm$63.4 &  52.8$\pm$88.8  & 72.8$\pm$84.7 & 65.1$\pm$82.8  & 44.7$\pm$98.8 & 20.8$\pm$98.0 & \textbf{113.8$\pm$26.3} & 87.1$\pm$62.5\\		
			\makecell[l]{TD3-TD3}  & 104.0$\pm$59.4 & 10.3$\pm$97.2 & 66.6$\pm$91.4 & 61.1$\pm$85.2  & 92.8$\pm$59.1 & 72.7$\pm$76.2 & 87.4$\pm$67.0 & 47.8$\pm$100.8\\
			%			\makecell{SAC-SAC} & -150.7$\pm$43.2 & -163.2$\pm$44.2 & -152.5$\pm$50.4 & -150.9$\pm$51.7 & -128.2$\pm$227.4  & -85.8$\pm$123.4& -41.0$\pm$117.6 & -104.5$\pm$94.4\\
			\hline
		\end{tabular}
	\end{center}
	%		\vskip -0.15in
	\vspace{-0.6cm}
\end{table*}

\subsection{Robustness Against the Changes in the Care-Receiver's Policy}
\begin{figure}
	\centering
	\includegraphics[width=0.8\columnwidth]{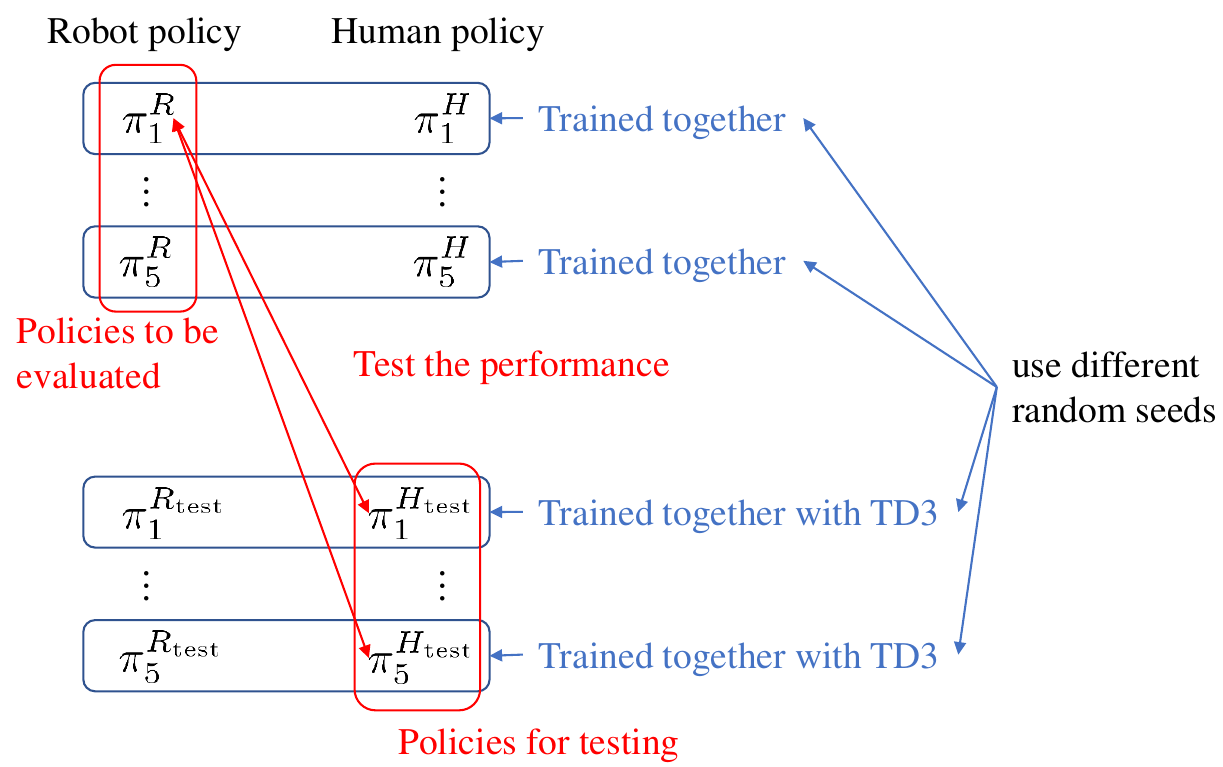}
	\caption{Procedure for evaluating the robustness of the caregiver's policy. A caregiver's policy was evaluated using a care-receiver's policy, which was trained separately.}
	\label{fig:eval_protocol}
	\vspace{-0.4cm}
\end{figure}

To investigate the robustness of the caregiver's policy against the change in the care-receiver's policy, we evaluated the performance of the caregiver's policy when it was used with the care-receiver's policy that was separately trained with TD3.
The evaluation procedure is summarized in Fig.~\ref{fig:eval_protocol}.
For PPO-LPPO-adv, PPO-LPPO and PPO-PPO, policies after training with 6 million steps were evaluated.
%For TD3-LTD3-adv and PPO-PPO, policies after training with 4 million steps were evaluated.
For TD3-TD3 and TD3-LTD3, policies after training with 2 million steps were evaluated because the policy performance got worse after 2 million steps.
We first train the caregiver's and care-receiver's policies using five different seeds with the proposed and baseline methods.
In the second step, we prepare another set of the caregiver's and care-receiver's policies trained using five different seeds with TD3-TD3. 
We then evaluate the performance of the caregiver's policy trained in the first step when it is used with the care-receiver's policy trained in the second step.
For comparison, we also show the performance of the caregiver's policy when it is used with the care-receiver's policy, which was trained together with it in the first step.
We performed 10 test episodes for each combination of policies, and report the average and standard deviation across five random seeds.

The results are shown in Table~\ref{tbl:td3_test}.
The column of ``train'' shows the policy performance when working with an agent trained together, and the column of ``test'' shows the performance when a caregiver's policy was used with a care-receiver's policy that was separately trained with TD3.
The bold text shows the best results in each task.
We examined the statistical difference based on unpaired t-test. When there are bold and non-bold numbers, it indicates that the there is a statistically significant difference between them, and p-value is less than 0.05. 
In baseline methods such as TD3-TD3 and PPO-PPO, the test performance was significantly lower than the training performance.
This result demonstrate that caregiver's policies obtained by a standard co-optimization are actually vulnerable to the change in the care-receiver's policies.
In contrast, the proposed method, PPO-LPPO-adv, clearly outperformed PPO-PPO in terms of the test performance, and the difference between the training and test performance was small in PPO-LPPO-adv.
This result demonstrate that the proposed method significantly improved the robustness of the caregiver's policy.
The difference between PPO-PPO and PPO-LPPO indicates the learning diverse behaviors improves the robustness of the caregiver's policy.
Furthermore, the difference between PPO-LPPO-adv and PPO-LPPO indicates the adversarial style sampling improves the robustness of the caregiver's policy.
The comparison between TD3-TD3, TD3-LTD3, and TD3-LTD3-adv aligns with this observation.
TD3-LTD3-adv outperformed TD3-LTD3, indicating the effectiveness of the proposed adversarial style sampling.
The learning of diverse care-receiver's behavior and adversarial style sampling improved the robustness of the caregiver's policy in TD3-based methods, whereas the PPO-based methods outperformed the TD3-based methods in our evaluation.

\section{Conclusions}
We presented a framework for robustifying a cooperative policy in multi-agent RL for assistive tasks.
%To transfer a policy from simulation to the real world, it is essential to address the robustness of caregiver's policy against the change in the care-receiver's policy.
In our framework, diverse care-receiver's responses are learned autonomously by maximizing the mutual information, 
and the caregiver's policy is robustified by generating care-receiver's responses in an adversarial manner during the training.
The proposed algorithm was evaluated in robotic assistive tasks implemented in Assistive Gym.
The experimental results showed that caregiver's policies obtained by standard co-optimization are vulnerable to the change in the care-receiver's policy.
% The results also revealed that the proposed framework can learn diverse strategies for the care-receiver.
The results also demonstrate that a caregiver's policy obtained by the proposed framework is more robust against changes in the care-receiver's policy.
%We believe that learning diverse behavior and adversarial training are promising approaches to improve the robustness of policies in cooperative multi-agent RL.
In future work, we will address challenges to be resolved to deploy a caregiver's policy in the real world.
Additionally, we plan to study other types of robustness in the future, such as those reported in \cite{He22,Bukharin23}.

\section*{ACKNOWLEDGMENT}
This work was partially supported by JST AIP Acceleration Research JPMJCR20U3, Moonshot R\&D Grant Number JPMJPS2011, CREST Grant Number JPMJCR2015, JSPS KAKENHI Grant Number JP19H01115, JP23K18476 and Basic Research Grant (Super AI) of Institute for AI and Beyond of the University of Tokyo.

%%%%%%%%%%%%%%%%%%%%%%%%%%%%%%%%%%%%%%%%%%%%%%%%%%%%%%%%%%%%%%%%%%%%%%%%%%%%%%%%

\bibliographystyle{IEEETran}
\bibliography{icra24}

\end{document}